\documentclass[10pt,twocolumn,letterpaper]{article}

\usepackage{cvpr}
\usepackage{times}
\usepackage{epsfig}
\usepackage{graphicx}
\usepackage{booktabs}
\usepackage{amsmath}
\usepackage{amssymb}
\usepackage{multicol}
\usepackage{multirow}
\usepackage{float}
\usepackage{multirow}
\usepackage{url}

\usepackage{makecell}
\usepackage{subcaption}
\usepackage{cuted}
\usepackage[font={small}]{caption}
\usepackage{pifont}
\usepackage[dvipsnames]{xcolor}
\usepackage{balance}
\usepackage{epigraph}

\usepackage{algorithm}
\usepackage{algorithmicx}
\usepackage{algpseudocode}

\definecolor{citecolor}{RGB}{65,105,225}
\usepackage[pagebackref=true,breaklinks=true,letterpaper=true,colorlinks,citecolor=citecolor,bookmarks=false]{hyperref}

\cvprfinalcopy 

\def\cvprPaperID{****} 

\begin{document}

\title{Everybody's Talkin': Let Me Talk as You Want}

\author{Linsen Song$^{1}\thanks{This work was done during an internship at SenseTime Research.}$\qquad Wayne Wu$^{2,3}$\qquad Chen Qian$^{2}$\qquad Ran He$^{1}$\qquad Chen Change Loy$^{3}$ \\
$^1$NLPR, CASIA \hspace{10pt} $^2$SenseTime Research \hspace{10pt} $^3$Nanyang Technological University \\
{\tt\small songlinsen2018@ia.ac.cn}\hspace{1cm}
{\tt\small \{wuwenyan, qianchen\}@sensetime.com}\hspace{1cm} \\
{\tt\small rhe@nlpr.ia.ac.cn}\hspace{1cm}
{\tt\small ccloy@ntu.edu.sg}
}

\maketitle

\begin{strip}
\centering
\vspace{-0.6in}
\includegraphics[width=\textwidth]{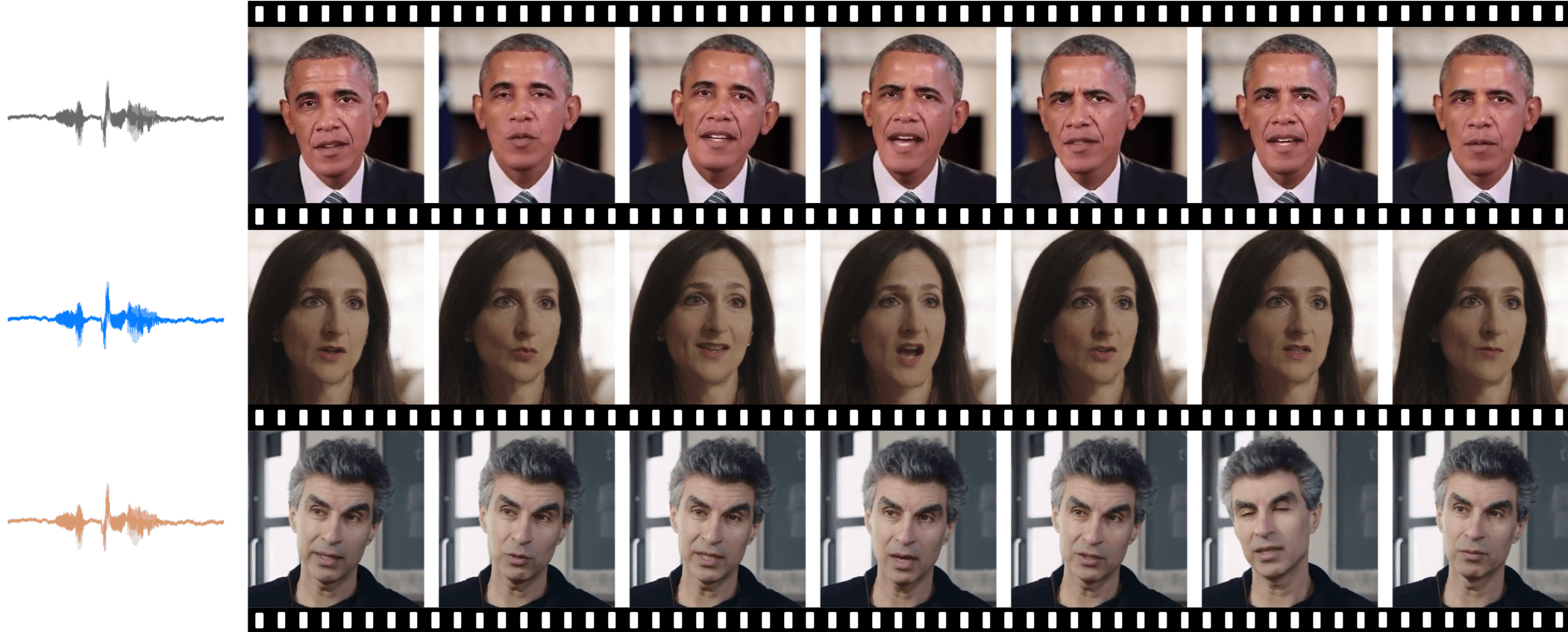}
\captionof{figure}{\textbf{Audio-based video editing.} Speech audio of an arbitrary speaker, extracted from any video, can be used to drive any videos featuring a random speaker.}
\label{fig:feature-graphic}
\end{strip}

\begin{abstract}

We present a method to edit a target portrait footage by taking a sequence of audio as input to synthesize a photo-realistic video.
This method is unique because it is highly dynamic. It does not assume a person-specific rendering network yet capable of translating arbitrary source audio into arbitrary video output.
Instead of learning a highly heterogeneous and nonlinear mapping from audio to the video directly, we first factorize each target video frame into orthogonal parameter spaces, i.e., expression, geometry, and pose, via monocular 3D face reconstruction. Next, a recurrent network is introduced to translate source audio into expression parameters that are primarily related to the audio content.
The audio-translated expression parameters are then used to synthesize a photo-realistic human subject in each video frame, with the movement of the mouth regions precisely mapped to the source audio. The geometry and pose parameters of the target human portrait are retained, therefore preserving the context of the original video footage.
%
%
Finally, we introduce a novel video rendering network and a dynamic programming method to construct a temporally coherent and photo-realistic video.
%
%
Extensive experiments demonstrate the superiority of our method over existing approaches. Our method is end-to-end learnable and robust to voice variations in the source audio. Some results are shown in Fig.~\ref{fig:feature-graphic}. Video results are shown on our project page\footnote[1]{Project Page: {\color{red}\url{https://wywu.github.io/projects/EBT/EBT.html}}}.

\end{abstract}

\section{Introduction}

\epigraph{I'm going where the sun keeps shining. Through the pouring rain. Going where the weather suits my clothes.}{\textit{Fred Neil, Everybody's Talkin'}}
\vspace{0.1cm}


Video portrait editing is a highly sought-after technique in view of its wide applications, such as filmmaking, video production, and telepresence. Commercial video editing applications, such as Adobe Premiere and Apple iMovie, are resource-intensive tools. Indeed, editing audio-visual content would require one or more footages to be reshot. Moreover, the quality of the edited video is highly dependent on the prowess of editors. 


Audio-based approach is an attractive technique for automatic video portrait editing.
Several methods~\cite{Chen2018LipMG, Zhu2018HighResolutionTF} are proposed to animate the mouth region of a still image to follow an audio speech. The result is an animated static image rather than a video, hence sacrificing realism. Audio-driven 3D head animation~\cite{Taylor2016AudiotoVisualSC} is an audio-based approach but aiming at a different goal, namely to drive stylized 3D computer graphic avatars, rather than to generate a photo-realistic video.
Suwajanakorn~\etal~\cite{Suwajanakorn2017SynthesizingOL} attempted to synthesize photo-realistic videos driven by audio. While impressive performance was achieved, the method assumes the source audio and target video to come from the same identity. The method is only demonstrated on the audio tracks and videos of Barack Obama. Besides, it requires long hours of single-identity data (up to 17 hours~\cite{Suwajanakorn2017SynthesizingOL}) for training using relatively controlled and high-quality shots.

In this paper, we investigate a learning-based framework that can perform many-to-many audio-to-video translation, \ie, without assuming a single identity of source audio and the target video. We further assume a scarce number of target video available for training, \eg, at most a 15-minute footage of a person is needed.
%
%
Such assumptions make our problem \textit{non-trivial}: 
1) Without sufficient data, especially in the absence of source video, it is challenging to learn direct mapping from audio to video. 
2) To apply the framework on arbitrary source audios and target videos, our method needs to cope with large audio-video variations between different subjects. 
3) Without explicitly specifying scene geometry, materials, lighting, and dynamics, as in the case of a standard graphics rendering engine, it is hard for a learning-based framework to generate photo-realistic yet temporally coherent videos.

To overcome the aforementioned challenges,  
we propose to use the expression parameter space, rather than the full pixels, as the target space for audio-to-video mapping.
This facilitates the learning of more effective mapping, since the expression is semantically more relevant to the audio source, compared to other orthogonal spaces, such as geometry and pose. In particular, we manipulate the expression of a target face by generating a new set of parameters through a novel LSTM-based Audio-to-Expression Translation Network. The newly generated expression parameters, combined with geometry and pose parameters of the target human portrait, allow us to reconstruct a 3D face mesh with the same identity and head pose of the target but with new expression (\ie, lip movements) that matches the phonemes of the source audio.

We further propose an Audio ID-Removing Network that keeps audio-to-expression translation agnostic to the identity of the source audio. Thus, the translation is robust to variations in the voices of different people in different source audio.
Finally, we solve the difficult face generation problem as a face completion problem conditioned on facial landmarks. Specifically, after reconstructing a 3D face mesh with new expression parameters, we extract the associated 2D landmarks from the mouth region and represent them as heatmaps. These heatmaps are combined with target frames where the mouth region is masked. Taking the landmark heatmaps and the masked target frames as inputs, a video rendering network is then used to complete the mouth region of each frame guided by dynamics of the landmarks.
%

We summarize our contributions as follows: 
1) We make the first attempt at formulating an end-to-end learnable framework that supports audio-based video portrait editing. We demonstrate coherent and photo-realistic results by focusing specifically on expression parameter space as the target space, from which source audios can be effectively translated into target videos.
2) We present an Audio ID-Removing Network that encourages an identity-agnostic audio-to-expression translation. This network allows our framework to cope with large variations in voices that are present in arbitrary audio sources.
3) We propose a Neural Video Rendering Network based on the notion of face completion with a masked face as input and mesh landmarks as conditions. This approach facilitates the generation of photo-realistic video for arbitrary people within one single network.

\noindent
\textbf{Ethical Considerations}.
Our method could contribute greatly towards advancement in video editing. We envisage relevant industries, such as filmmaking, video production, and telepresence to benefit immensely from this technique. We do acknowledge the potential of such forward-looking technology being misused or abused for various malevolent purposes, \eg, aspersion, media manipulation, or dissemination of malicious propaganda. Therefore, we strongly advocate and support all safeguarding measures against such exploitative practices. We welcome enactment and enforcement of legislation to obligate all edited videos to be distinctly labeled as such, to mandate informed consent be collected from all subjects involved in the edited video, as well as to impose hefty levy on all law defaulters. Being at the forefront of developing creative and innovative technologies, we strive to develop methodologies to detect edited video as a countermeasure. We also encourage the public to serve as sentinels in reporting any suspicious-looking videos to the authority. Working in concert, we shall be able to promote cutting-edge and innovative technologies without compromising the personal interest of the general public.

\section{Related Work}

\noindent
\textbf{Audio-based Facial Animation.}
Driving a facial animation of a target 3D head model by input source audio learns to associate phonemes or speech features of source audio with visemes. Taylor \etal~\cite{Taylor2017ADL} propose to directly map phonemes of source audio to face rig. Afterward, many speech-driven methods have been shown superior to phoneme-driven methods under different 3D models, \eg face rig~\cite{Zhou2018Visemenet, Edwards2016JALIAA}, face mesh~\cite{Karras2017AudiodrivenFA}, and expression blendshapes~\cite{Pham2017SpeechDriven3F}.

Compared to driving a 3D head model, driving a photo-realistic portrait video is much harder since speaker-specific appearance and head pose are crucial for the quality of the generated portrait video. Taylor \etal~\cite{Taylor2016AudiotoVisualSC} present a sliding window neural network that maps speech feature window to visual feature window encoded by active appearance model (AAM)~\cite{Cootes1998ActiveAM} parameters. 
To improve the visual quality, in several methods~\cite{Jamaludin2019YouST, Vougioukas2018EndtoEndSF, Zhou2018TalkingFG}, a still face image is taken as a reference for video generation. However, the result is an animation of a still image rather than a natural video.

Recently, Suwajanakornet \etal~\cite{Suwajanakorn2017SynthesizingOL} obtain the state-of-the-art result in synthesizing the Obama video portrait. However, it assumes the source and target to have the same identity and requires long hour of training data (up to 17 hours). Thus, it is not applicable in audio-based video editing that need to cope with different sources of voice and target actors. In addition, the target video data is relatively scarce. Fried \etal~\cite{Fried2019TextbasedEO} proposed a method to edit a talking-head video based on its transcript to produce a realistic video. While it produce compelling results, a person-specific face rendering network need to be trained for each target person. Besides, it takes a long time for viseme search (up to 2 hours for 1-hour recording) and relies on phoneme, thus it cannot be scaled to different languages.

\noindent
\textbf{Video-based Facial Reenactment.} It is inherently difficult to synthesize mouth movements based solely on speech audio. Therefore, many methods turn to learning mouth movements from videos comprising the same/intended speech content.~\cite{Geng2018WarpguidedGF, Wiles2018X2FaceAN, Thies2016Face2FaceRF, Kim2018DeepVP, Nagano2018paGANRA}. From source portrait video, facial landmarks~\cite{Geng2018WarpguidedGF,shengju2019makeaface} or expression parameters~\cite{Wiles2018X2FaceAN} are estimated to drive the target face image. In all of these methods~\cite{Geng2018WarpguidedGF, shengju2019makeaface, Wiles2018X2FaceAN}, the generated portrait videos are frame-wise realistic but they suffer from poor temporal continuity. ReenactGAN~\cite{wayne2018reenactgan} is the first end-to-end learnable video-based facial reenactment method. It introduces a notion of ``boundary latent space'' to perform many-to-one face reenactment. However, ReenactGAN needs person-specific transformers and decoders, which makes the model size increase linearly with persion identities raising.

Many model-based methods~\cite{Thies2016Face2FaceRF, Kim2018DeepVP, Nagano2018paGANRA} leverage a 3D head model to disentangle facial geometry, expression, and pose. Face2Face~\cite{Thies2016Face2FaceRF} transfers expressions in parameter space from the source to the target actor. To synthesize a realistic target mouth region, the best mouth image of the target actor is retrieved and warped~\cite{Thies2016Face2FaceRF}. Kim \etal~\cite{Kim2018DeepVP} present a method that transfers expression, pose parameters, and eye movement from the source to the target actor. These methods are person-specific~\cite{Kim2018DeepVP} therefore rigid in practice and suffer audio-visual misalignment~\cite{Thies2016Face2FaceRF, Kim2018DeepVP, Nagano2018paGANRA} therefore creating artifacts leading to unrealistic results.

\noindent
\textbf{Deep Generative Models.}
Inspired by the successful application of GAN~\cite{Goodfellow2014GenerativeAN} in image generation~\cite{Radford2015UnsupervisedRL, Mao2016LeastSG, Zhu2017UnpairedIT, Huang2017BeyondFR,wayne2019transgaga}, many methods~\cite{Chen2018LipMG, Zhou2018TalkingFG, Zhu2018HighResolutionTF, Pham2018GenerativeAT, Kim2018DeepVP} leverage GAN to generate photo-realistic talking face images conditioned on coarse rendering image~\cite{Kim2018DeepVP}, fused audio, and image features~\cite{Chen2018LipMG, Zhu2018HighResolutionTF, Zhou2018TalkingFG}. Generative inpainting networks~\cite{Pathak2016ContextEF, Iizuka2017GloballyAL, Liu2018ImageIF, Yu2018FreeFormII} are capable of modifying image content by imposing guided object edges or semantic maps~\cite{Yu2018FreeFormII, Song2018GeometryAwareFC}. We convert talking face generation into an inpainting problem of mouth region, since mouth movement is primarily induced by input speeches.

\noindent
\textbf{Monocular 3D Face Reconstruction.} Reconstructing 3D face shape and texture from a single face image has extensive applications in face image manipulation and animation~\cite{Fyffe2014DrivingHF, Garrido2015VDubMF, Thies2016Face2FaceRF, Roth2016Adaptive3D,  Suwajanakorn2017SynthesizingOL}. In general, monocular 3D face reconstruction produces facial shape, expression, texture, and pose parameters by solving a non-linear optimization problem constrained by a statistical linear model of facial shape and texture, such as Basel face model~\cite{Paysan2009A3F}, FaceWarehouse model~\cite{Cao2014FaceWarehouseA3}, and Flame~\cite{Li2017LearningAM}. Recently, different 3D head models have been increasingly applied in talking portrait video synthesis~\cite{Thies2016Face2FaceRF, Suwajanakorn2017SynthesizingOL, Kim2018DeepVP, Nagano2018paGANRA, Fried2019TextbasedEO, Kim2019NeuralSV}.

\begin{figure*}[h]
\centering
\includegraphics[width=1.0\linewidth]{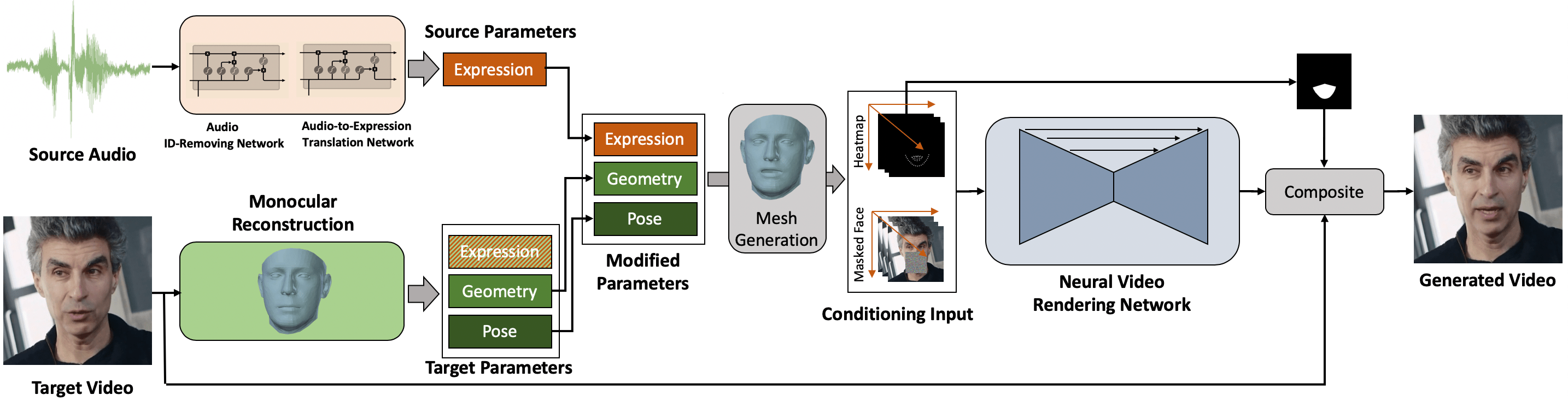}
\vspace{-5mm}
\caption{\textbf{Architecture.} Our network contains an Audio-to-Expression Translation Network that learns facial expression parameters from speech audio and a Neural Video Rendering Network that generates mouth region guided by projected mouth landmarks.}
\vspace{-5mm}
\label{fig:fig_architecture}
\end{figure*}


\section{Methodology}

The architecture of the proposed method is shown in Fig.~\ref{fig:fig_architecture}. First, we register a parametric 3D face model~\cite{Cao2014FaceWarehouseA3} in the target video, for every portrait video frame to extract face geometry, pose, and expression parameters. Then, the Audio-to-Expression Translation Network learns the mapping from the source audio feature to face expression parameters. We design an Audio ID-Removing Network to alleviate the issues on large variations caused by multiple speakers. Lastly, we formulate the talking face generation problem as a face completion problem guided by mouth region landmarks, in which the landmarks are projected from the restructured 3D facial mesh. We propose a Neural Video Rendering network to complete the mouth region of each frame, guided by the dynamics of the landmarks to generate a photo-realistic portrait video.

\subsection{3D Face Modeling}


We leverage a parametric 3D face model~\cite{Cao2014FaceWarehouseA3} on portrait video frame to recover low dimensional geometry, expression, and pose parameters. To reduce parameter dimension, geometry and expression bases are computed based on high-quality head scans~\cite{Blanz1999AMM} and facial blendshapes~\cite{Cao2014FaceWarehouseA3, Alexander2010TheDE} via principal component analysis (PCA). The geometry parameters $s\in \mathbb{R}^{199}$ and the expression parameters $e\in \mathbb{R}^{29}$ are the coefficients of geometry and expression principle components in the PCA, respectively. The pose of the head $p\in \mathbb{R}^{6}$ which contains 3 head rotation coefficients, 2 translation coefficients ($x$ and $y$ directions on the screen surface), and 1 scaling coefficient. All the parameters are computed by solving a non-linear optimization problem, constrained by the statistical linear 3D face model~\cite{Blanz1999AMM}. 
By optimizing the geometry, expression, and pose parameters of a given monocular face image based on its detected facial landmarks, protrait video frames will be automatically annotated with low dimensional vectors~\cite{Blanz1999AMM}. The recovered expression parameters are used as the learning target in the Audio-to-Expression Translation Network. Then, the recovered geometry and pose parameters, together with the expression parameters inferred by the Audio-to-Expression Translation Network, are employed for reconstructing the 3D facial mesh. 

\subsection{Audio-to-Expression Translation}


\subsubsection{Audio ID-Removing Network}

We empirically find that identity information embedded in the speech feature degrades the performance of mapping speech to mouth movement. Inspired by recent advances of the speaker adaptation method in the literature of speech recognition~\cite{Visweswariah2002ST, Povey2012AB}, we transfer the speech feature lies in different speaker domains onto a ``global speaker'' domain by applying a linear transformation, in the form~\cite{Visweswariah2002ST}:

\begin{equation}
\begin{array}{c}
x^{'}=W_i x + b_i = \bar{W}_i \bar{x}, \\
{\rm where}\ \bar{W_i}=(W_i,\ b_i),\ \bar{x}=(x;\ 1),\\
\bar{W_i}=I+\sum_{j=1}^{k} \lambda_j \bar{W}^j
\end{array}
\label{eq_audio}
\end{equation}
Here, $x$ and $x^{'}$ represent the raw and transferred speech feature, respectively, while $\bar{W_i}=I+\sum_{j=1}^{k} \lambda_j \bar{W}^j$ represents the speaker-specific adaptation parameter that is factorized into an identity matrix $I$ plus weighted sum of $k$ components $\bar{W}^j$~\cite{Povey2012AB}. In speech recognition, these parameters are iteratively optimized by fMLLR~\cite{Digalakis1995SA, Gales1998ML} and EM algorithms. We formulate the above method into a neural network to be integrated with our end-to-end deep learning network.

\begin{figure}[t]
\centering
\includegraphics[width=1.0\linewidth]{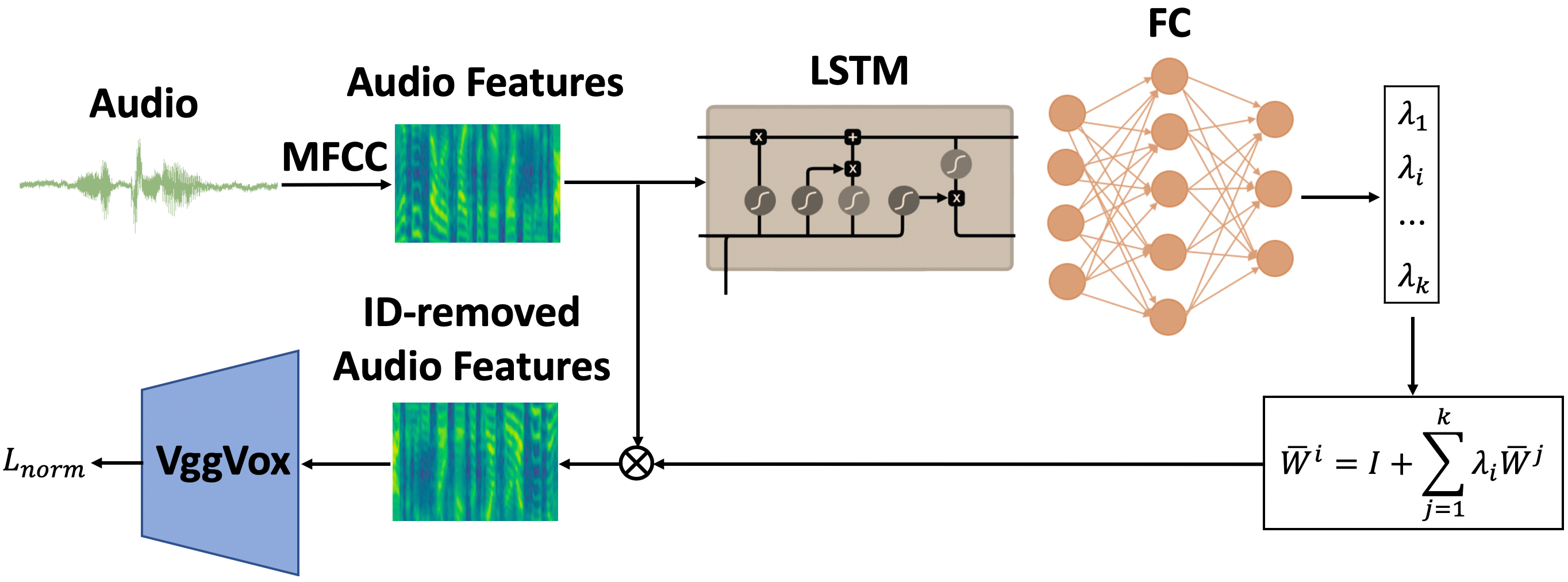}
\caption{\textbf{Audio ID-Removing Network.} We formulate the speaker adaptation method from speech recognition~\cite{Visweswariah2002ST, Povey2012AB} as a neural network. The network removes identity in speech MFCC spectrum by transferring it to the ``global speaker'' domain.}
\vspace{-5mm}
\label{fig_id_remove}
\end{figure}

From Eq.(\ref{eq_audio}), the parameters $\lambda_j$ need to be learned from the input speech feature, while the matrix components $\bar{W}^j\ $ is general speech features of different speakers. Thus, we design an LSTM+FC network to infer $\lambda_j$ from the input and set the matrix components $\bar{W}^j$ as the optimizing parameter of the Audio ID-Removing Network. The matrix components $\bar{W}^j$ of the Audio ID-Removing Network are updated by the gradient descent-based algorithm. The details of the network is depicted in Fig.~\ref{fig_id_remove}. The output of the Audio ID-Removing Network is a new MFCC (Mel-frequency cepstral coefficients) spectrum.  We apply a pre-trained speaker identity network VGGVox~\cite{Nagrani2017VA, Nagrani2017VD} on the new MFCC spectrum and constrain the Audio ID-Removing Network by the following cross-entropy loss function:

\begin{equation}
\mathcal{L}_{norm} = -\sum_{c=1}^{N} \frac{1}{N}\log p(c|x^{'}),
\label{eq_vggvox}
\end{equation}

\noindent
where $N$ is the number of speakers, $c$ is the speaker class label. The $p(c|x^{'})$ is the probability of assigning MFCC $x^{'}$ to speaker $c$, which is inferred from the pre-trained VGGVox. Eq.~\eqref{eq_vggvox} enforces the Audio ID-Removing Network to produce an MFCC spectrum that is not distinguishable by the pre-trained VGGVox.

\subsubsection{Audio-to-Expression Translation Network}


We formulate a simple but effective Audio-to-Expression Translation Network that learns the mapping from the ID-removed MFCC feature to the corresponding facial expression parameters. To infer the expression parameters at time $t$, the translation network observes a sliding window speech clip of 1 second, which contains 0.8 seconds before time $t$ and 0.2 seconds after time $t$.

We empirically find it challenging to train a network to solely regress the expression parameters. The underlying reason could be that the expression parameters are defined and related to the 3DMM model that is hard to model by the network.
To facilitate the learning, we introduce a shape constraint.
In particular, with the predicted expression parameters from audio and the ground truth geometry/pose parameters of the video portrait, we can obtain a predicted reconstructed 3D facial mesh. Then, we project 3D points of mouth area to the 2D space to obtain the predicted 2D mouth landmarks. Using a similar method, we can obtain a set of ground-truth 2D mouth landmarks from the ground-truth expression parameters. The shape constraint can be introduced between the predicted 2D mouth landmarks and ground-truth 2D mouth landmarks.
The whole process of generating mouth landmarks from expression parameters only involves linear operations and thus is differential. The loss function is written as follows:
%
\begin{equation}
\mathcal{L}_{trans}=\mathcal{L}_{exp}+\mathcal{L}_{shape}=||\hat{e}-e||_2+||\hat{l}-l||_2,
\label{eq_audio2expldmk}
\end{equation}
%
where $e$ and $l$ are the ground truth expression and landmark, respectively, and $\hat{e}$ and $\hat{l}$ are the output expression and landmark of the translation network, respectively. The Audio ID-Removing and Audio-to-Expression Translation Networks are trained jointly, whose objective function is weighted sum of $L_{norm}$ (Eq.~\eqref{eq_vggvox}) and $L_{trans}$ (Eq.~\eqref{eq_audio2expldmk}).

\subsection{Neural Video Rendering Network}

\subsubsection{Network Architecture}

Our final step is to generate photo-realistic talking face video that is conditioned on dynamic background portrait video and is guided by the mouth region landmark heatmap sequence. We design a completion-based generation network that completes the mouth region guided by mouth landmarks. 
First, to obtain the masked face images, a tailored dynamic programming based on retiming algorithm inspired by ~\cite{Suwajanakorn2017SynthesizingOL} is introduced to select frame sequence whose head shaking and blink of eyes look compatible with the source speech. Then, the mouth area that contains lip, jaw, and nasolabial folds are manually occluded by a square mask filled with random noise. To make the conversion from the landmark coordinates to heatmap differentiable, we follow~\cite{Jakab2018conditional,wayne2019disentangling} to generate heatmaps with Gaussian-like functions centered at landmark locations.
We modify a Unet~\cite{Ronneberger2015UNetCN,shengju2019makeaface}-based network as our generation network. The employed skip-connection enables our network to transfer fine-scale structure information. In this way, the landmark heatmap at the input can directly guide the mouth region generation at the output, and the structure of the generated mouth obeys the heatmaps~\cite{shengju2019makeaface, Wang2019ExampleGuidedSC}. 

We composite the generated mouth region over the target face frame according to the input mouth region mask. To obtain the mouth region mask, we connect the outermost mouth landmarks as a polygon and fill it with white color, then we erode the binary mask and smooth its boundaries with a Gaussian filter~\cite{Kim2019NeuralSV}. With the soft mask, we leverage Poisson blending~\cite{Prez2003PoissonIE} to achieve seamless blending. To improve the temporal continuity of generated video, we apply a sliding window on the input masked video frames and heatmaps~\cite{Kim2018DeepVP, Kim2019NeuralSV}. The input of the Neural Video Rendering Network is a tensor stacked by 7 RGB frames and 7 heatmap gray images~\cite{Kim2019NeuralSV}. It works well in most cases while a little lip motion jitters and appearance flicker might emerge in the final video. Then, a video temporal flicker removal algorithm improved from ~\cite{Bonneel2015BlindVT} is applied to eliminate these artifacts. Please refer to \textit{appendix} for more details of the flicker removal algorithm.


\subsubsection{Loss Functions}

The loss function for training the Neural Video Rendering Network is written as follows:
\vspace{-1mm}
\begin{equation}
\mathcal{L}_{render} = \mathcal{L}_{recon} + \mathcal{L}_{adv} + \mathcal{L}_{vgg} + \mathcal{L}_{tv} + \mathcal{L}_{gp}.
\label{eq_render}
\end{equation}
\vspace{-5mm}

\noindent
The reconstruction loss $L_{recon}$ is the pixel-wise L1 loss between the ground truth and generated images. To improve the realism of the generated video, we apply the LSGAN~\cite{Mao2017LS} adversarial loss $L_{adv}$ and add the gradient penalty term $L_{gp}$~\cite{Gulrajani2017IT} for faster and more stable training. We also apply the perception loss $L_{vgg}$~\cite{Johnson2016PL} to improve the quality of generated images by constraining the image features at different scales. The total variation regularization term $L_{tv}$ is used to reduce spike artifact that usually occurs when $L_{vgg}$ is applied~\cite{Johnson2016PL}. The network is trained end-to-end with $L_{total} = L_{norm} + L_{trans} + L_{render}$ (Equations~\eqref{eq_vggvox},\eqref{eq_audio2expldmk}, and \eqref{eq_render}) with different coefficients. Due to the limited space, we report the details of the loss function, network architecture, and experimental settings in our \textit{appendix}.

\begin{figure*}[ht!]
    \centering
    \includegraphics[width=0.9\linewidth]{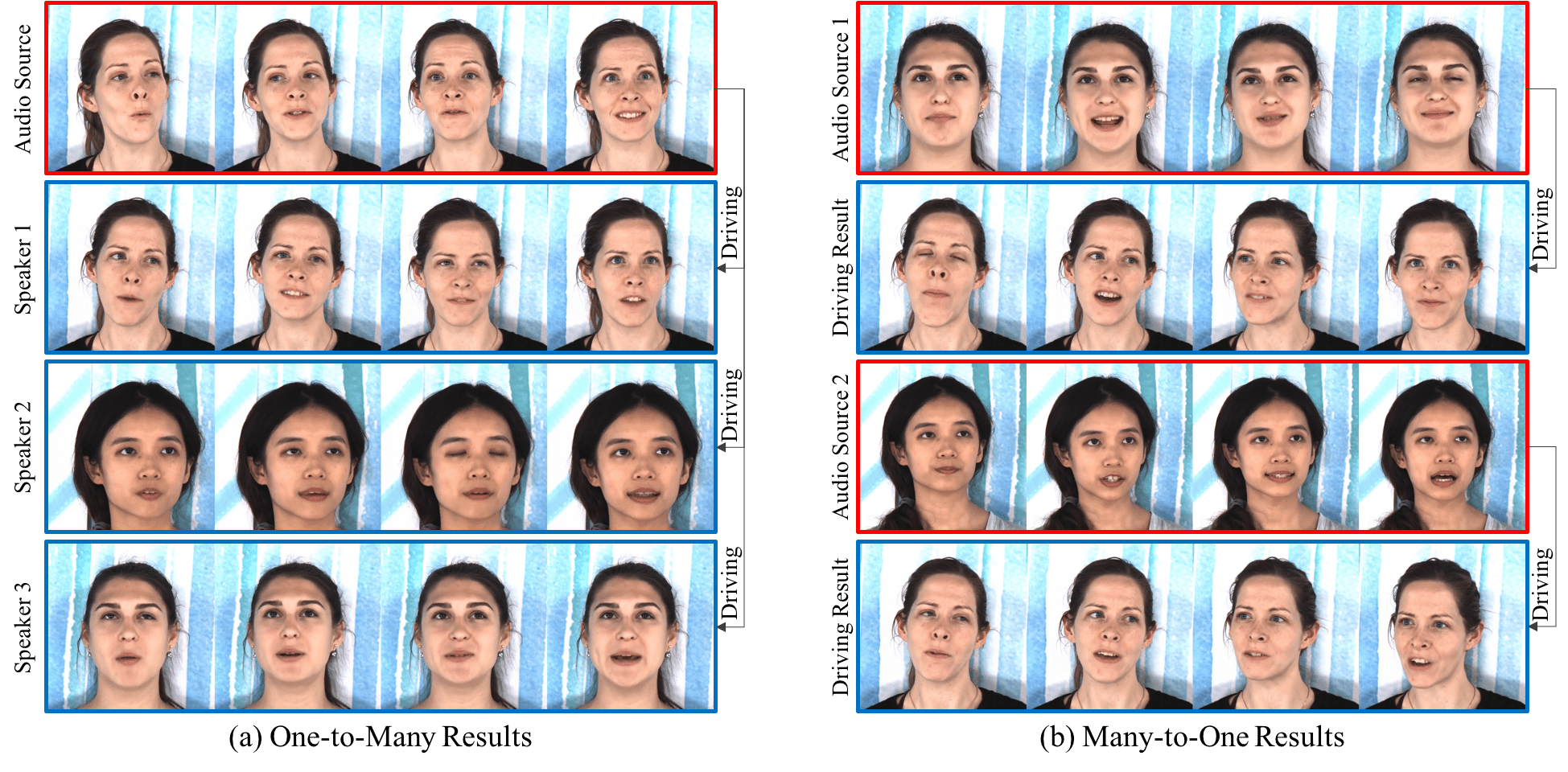}
    \vspace{-5mm}
    \caption{\textbf{Many-to-many results.} (a) One-to-many results: we use speech audio of $1$ speaker to drive face of $3$ different speakers. (b) Many-to-one results: we use speech audio of $2$ different speakers to drive $1$ same speaker. For video results we refer to the video on our project page.}
    \vspace{-8mm}
    \label{fig_many_to_many}
\end{figure*}

\section{Results}

We show qualitative and quantitative results on a variety of videos to demonstrate the superiority of our method over existing techniques and the effectiveness of proposed components. 

\noindent
\textbf{Datasets.} We evaluate our method on a talking face benchmark dataset GRID~\cite{Cooke2006AA} and a speech video dataset we newly collected. The former contains 1,000 sentences spoken by 18 males and 16 females. We follow Chen~\etal~\cite{Chen2018LipMG} to split training and testing sets on the GRID dataset. Since GRID only provides frontal face videos of minor head movement, we record a video dataset that contains multiple head poses and time-varying head motion. The collected dataset contains speech videos of 4 speakers. Each speaker contributes 15 minutes video for training and 2 minutes video for testing, all videos are captured from 7 viewpoints to provide 7 head poses. Resolution of each video is $1920\times1080$. We also take several videos downloaded from YouTube with the same percentage of the training and testing split of recorded data to evaluate our approach.

\noindent
\textbf{Evaluation Metrics.}
To evaluate the accuracy of the expression parameters and the projected landmarks under various head poses and motions, we apply the following distance metric:

\vspace{-5mm}
\begin{equation}
\begin{array}{c}
E_{exp} = \frac{1}{N_{exp}}\sum_{i=1}^{N_{exp}}||\widehat{e}(i)-e(i)||_2,\\
E_{ldmk} = \frac{1}{N_{ldmk}}\sum_{i=1}^{N_{ldmk}}||\widehat{l}(i)-{l}(i)||_2,
\end{array}
\label{eq_exp_ldmk_err}
\end{equation}
\vspace{-2mm}

\noindent
where $N_{ldmk}$ and $N_{exp}$ are the number of landmarks and expression parameters respectively and $i$ is the index of landmarks or expression parameters. To quantitatively evaluate the generated quality of portrait videos, we apply common image quality metrics like PSNR~\cite{Wang2004IQ} and SSIM~\cite{Wang2004IQ}. To qualitatively evaluate the generated quality of portrait videos, we conduct a user study in Section~\ref{section_user_study} and demonstrate some generated video results on our project page.


\subsection{Audio-to-Video Translation}

\noindent
\textbf{Many-to-Many Results.} To prove that the audio-to-expression network is capable of handling various speakers and the face completion network is generalized on multiple speakers, we present one-to-many results and many-to-one results in Fig.~\ref{fig_many_to_many} and on our project page. In the one-to-many results, we use the speech audio of one speaker to drive different speakers. Note that different speakers share a single generator instead of multiple person-specific generators. In the many-to-one results, we use the speech audio of different speakers to drive the same speaker. This is in contrast to recent methods, where the whole pipeline~\cite{Suwajanakorn2017SynthesizingOL} or part of components~\cite{Kim2019NeuralSV, Fried2019TextbasedEO} is designed for a specific person, which disables these methods in handling different voice timbres and facial appearances.

\begin{figure}[h!]
    \centering
    \includegraphics[width=1\linewidth]{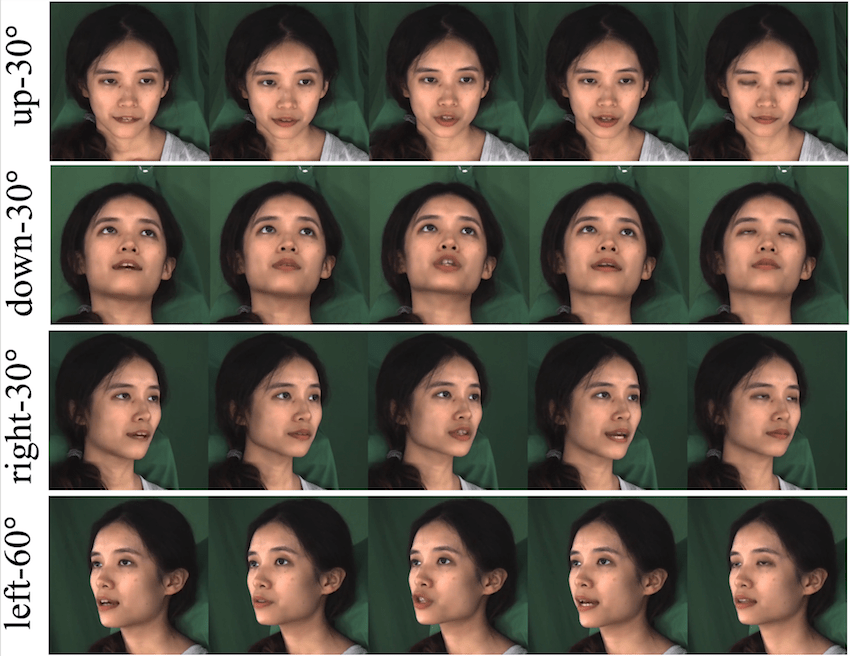}
    \vspace{-5mm}
    \caption{\textbf{Large pose results.} We demonstrate $4$ head poses including up, down, right and left. Video results of all of $7$ head poses can be viewed the video on our project page.}
    \vspace{-5mm}
    \label{fig_large_pose}
\end{figure}

\noindent
\textbf{Large Pose Results.} The main purpose of leveraging 3D face model is to handle head pose variations in generating talking face videos. As far as we know, majority of the recent audio-driving methods focus on generating frontal face video no matter whether a 3D head model is applied~\cite{Suwajanakorn2017SynthesizingOL, Kim2019NeuralSV} or not~\cite{Vondrick2016GeneratingV, Zhou2018Visemenet, DBLP:conf/cvpr/ChenMDX19}. Our method, however, can generate portrait videos under various large poses driven by audio input. Thanks to the decomposition of audio-driving facial animation problem in our framework, which makes the audio only relate to expression parameters of face rather than shape or pose parameters.  Results are shown in Fig.~\ref{fig_large_pose} and the video on our project page. Note that in previous methods~\cite{Suwajanakorn2017SynthesizingOL,Zhou2018Visemenet,DBLP:conf/cvpr/ChenMDX19}, they directly learn a mapping from audio to landmarks, which involves the shape and pose information that is actually independent to the input audio.

\begin{figure}[h!]
    \centering
    \includegraphics[width=1\linewidth]{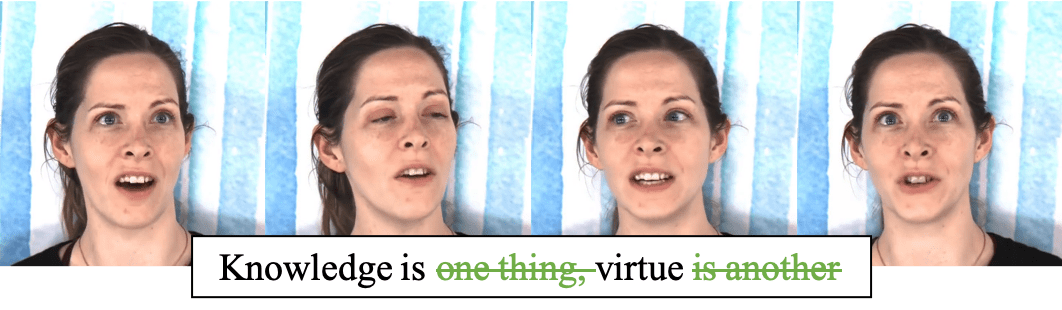}
    \vspace{-5mm}
    \caption{\textbf{Audio editing.} We select ``knowledge is'' and ``virtue'' from ``Knowledge is one thing, virtue is another'' in the source audio, then recombine them as ``Knowledge is virtue'' as input.}
    \vspace{-5mm}
    \label{fig_audio_edit}
\end{figure}

\begin{figure}[h]
    \centering
    \includegraphics[width=1\linewidth]{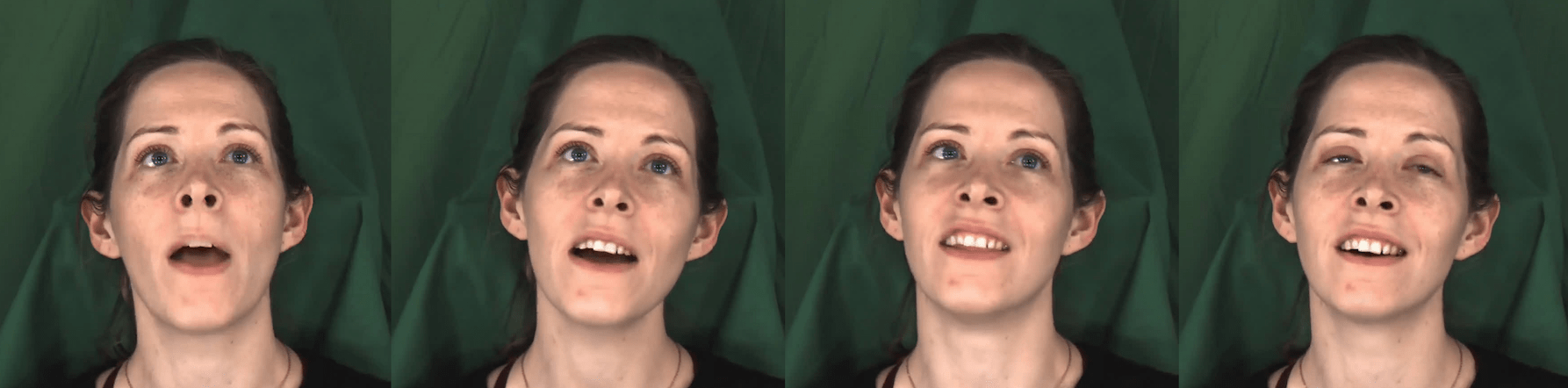}
    \vspace{-5mm}
    \caption{\textbf{Singing.} We evaluate our network on the singing audio clips. Video result can be viewed in the video on our project page.}
    \vspace{-5mm}
    \label{fig_sing}
\end{figure}

\noindent
\textbf{Audio Editing \& Singing Results.} Our method can also be used to edit the speech contents of a pre-recorded video by splitting and recombining the words or sentences taken from any source audio. We show our audio editing results in Fig.~\ref{fig_audio_edit} and video on our project page. In addition, we also ask a person to record singing and the audio is fed into our network. The driving result can be viewed in Fig.~\ref{fig_sing} and video on our project page. This demonstrates the generalization capability of our method and its potential in more complex audio-to-video tasks.

\subsection{Comparison with State-of-the-Art}
\label{sec_comparison_sota}

\begin{figure}[ht!]
    \centering
    \includegraphics[width=1\linewidth]{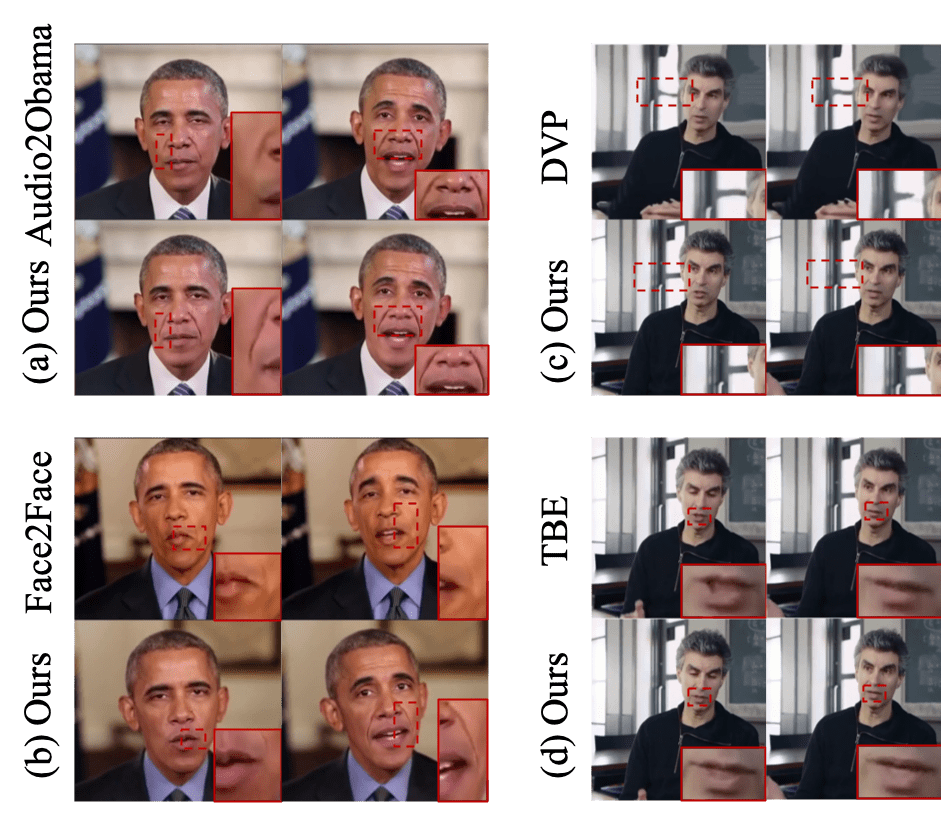}
    \vspace{-5mm}
    \caption{\textbf{Comparison to state-of-the-art methods.} Comparison between our method with Audio2Obama~\cite{Suwajanakorn2017SynthesizingOL}, Face2Face~\cite{Thies2016Face2FaceRF}, DVP~\cite{Kim2018DeepVP}, and TBE~\cite{Fried2019TextbasedEO}.}
    \vspace{-5mm}
    \label{fig_qualitative_compare}
\end{figure}

We compare our method with the recent state-of-the-art portrait video generation methods, \eg, Audio2Obama~\cite{Suwajanakorn2017SynthesizingOL}, Face2Face~\cite{Thies2016Face2FaceRF}, DVP (Deep Video Portrait)~\cite{Kim2018DeepVP} and Text-based Editing (TBE)~\cite{Fried2019TextbasedEO}. The comparative results are demonstrated in Fig.~\ref{fig_qualitative_compare} and video on our project page.

First, the Audio2Obama~\cite{Suwajanakorn2017SynthesizingOL} combines a weighted median texture for synthesizing lower face texture and a teeth proxy for capturing teeth sharp details. Our GAN-based rendering network generates better texture details compare to the weighted median texture synthesis~\cite{Suwajanakorn2017SynthesizingOL}, \eg, nasolabial folds (Fig.~\ref{fig_qualitative_compare} (a)). Then, we compare our method to Face2Face~\cite{Thies2016Face2FaceRF} that supports talking face generation driving by source video in Fig.~\ref{fig_qualitative_compare} (b). Face2Face~\cite{Thies2016Face2FaceRF} directly transfers facial expression of source video in the parameter space while our method infers facial expression from source audio. The similar lip movement of Face2Face and our method in Fig.~\ref{fig_qualitative_compare} (b) suggests the effectiveness of our Audio-to-Expression Translation Network in learning accurate lip movement from speech audio. Moreover, our GAN-based rendering network generates better texture details, such as mouth corners and nasolabial folds. We also compare to another video-driving method DVP~\cite{Kim2018DeepVP} that supports talking face generation (Fig.~\ref{fig_qualitative_compare} (c)). In DVP, a rendering-to-video translation network is designed to synthesize the whole frame other than the face region. It avoids the blending of face region and background that might be easily detectable. The DVP might fail in a complex and dynamic background as shown in Fig.~\ref{fig_qualitative_compare} (c). In contrast, our method uses the original background and achieves seamless blending that is hard to distinguish. Finally, we compare our method with the contemporary text-based talking face editing method TBE~\cite{Fried2019TextbasedEO} in Fig.~\ref{fig_qualitative_compare} (d). In TBE, the mouth region is searched by phoneme and a semi-parametric inpainting network is proposed to inpaint the seam between the retrieved mouth and the original face background. This method requires training of a person-specific network per input video while our method can generalize on multiple speakers and head poses. Besides, our generation network produces competitive mouth details as shown in Fig.~\ref{fig_qualitative_compare} (d).

\subsection{Ablation Study}
\begin{figure}[h]
    \centering
    \includegraphics[width=0.9\linewidth]{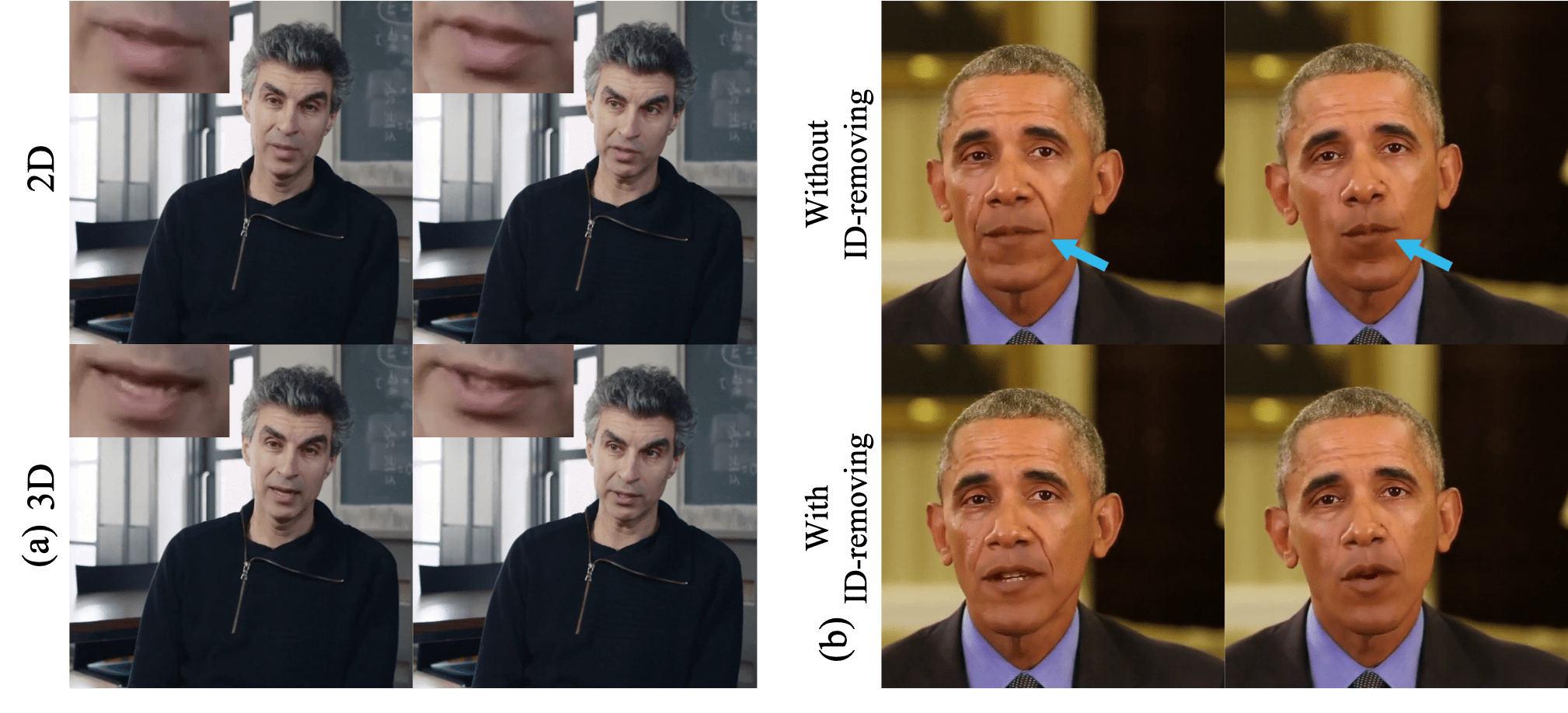}
    \vspace{-4mm}
    \caption{\textbf{(a) 2D vs 3D qualitative comparison.} 3D parameter regression outperforms 2D parameter regression for head motion and poses. \textbf{(b) ID-removing qualitative comparison.} Improvement can be observed on a failure case caused by not applying Audio ID-Removing Network.}
    \vspace{-3mm}
    \label{fig:abl_2d_3d_id_remove}
\end{figure}

\noindent
\textbf{Evaluation of Parameter Regression.} To prove the superiority of incorporating the 3D face model, we compare our network with the one that replaces Audio-to-Expression Translation Network with an Audio-to-Landmark Translation Network as performed in~\cite{Suwajanakorn2017SynthesizingOL}. The Audio-to-Landmark Translation Network modifies the last fully connected layer of the Audio-to-Expression Translation Network so that its output dimension is the coordinate number of mouth region landmarks. The visualized comparison can be viewed in the video on our project page and Fig.~\ref{fig:abl_2d_3d_id_remove} (a). We also compare the quantitative metric on GRID and collected dataset as shown in Tab.~\ref{tab_2d_3d}. In the collected dataset that contains more head motion and poses, our method achieves better lip synchronization results as the mouth generated by the one that applies Audio-to-Landmark Translation Network does not even open.

\begin{table}[h]
    \centering
    \vspace{-2mm}
    \caption{\textbf{2D vs 3D quantitative comparison.} $E_{exp}$, $E_{ldmk}$, PSNR, and SSIM comparison of 2D and 3D parameter regression.}
    \resizebox{1.0\linewidth}{!}{
    \small
    \begin{tabular}{c|c|cc|cc}
    \Xhline{1.2pt}
    \multicolumn{2}{c|}{Parameter Regression} & $E_{exp}$ & $E_{ldmk}$ & PSNR & SSIM \\
    \Xhline{1.2pt}
    \multirow{2}{*}{GRID} & 2D & - & 3.99 & 28.06 & 0.89 \\
    & \textbf{3D} & \textbf{0.65} & \textbf{2.24} & \textbf{31.19} & \textbf{0.95} \\
    \Xhline{1.2pt}
    \multirow{2}{*}{Collected} & 2D & - & 3.13 & 26.76 & 0.93 \\
    & \textbf{3D} & \textbf{0.595} & \textbf{1.82} & \textbf{29.16} & \textbf{0.95} \\
    \Xhline{1.2pt}
    \end{tabular}
    \label{tab_2d_3d}}
    \vspace{-2mm}
\end{table}

\begin{table}[h]
    \caption{\label{tab_abl} \textbf{ID-removing quantitative comparison.} ``BL'' is the 3D parameter regression baseline; ``IR'' is ``id-removing''.}
    \begin{center}
    \resizebox{1.0\linewidth}{!}{
    \begin{tabular}{c|c|cc|cc}
    \Xhline{1.2pt}
    \multicolumn{2}{c|}{Method} & $E_{exp}$ & $E_{ldmk}$ & PSNR & SSIM \\
    \Xhline{1.2pt}
    \multirow{2}{*}{GRID} & BL & 0.84 & 3.09 & 30.88 & 0.94 \\
    & +IR & \textbf{0.65} & \textbf{2.24} & \textbf{32.23} & \textbf{0.97} \\
    \Xhline{1.2pt}
    \multirow{2}{*}{Collected} & BL & 0.68 & 1.92 & 26.86 & 0.95 \\
    & +IR & \textbf{0.59} & \textbf{1.83} & \textbf{31.21} & \textbf{0.96} \\
    \Xhline{1.2pt}
    \end{tabular}
    }
    \end{center}
\end{table}

\noindent
\textbf{Evaluation of ID-removing.} Our Audio ID-Removing Network transfers the speech feature of different speakers to a ``global speaker'', which can be directly proven by tSNE~\cite{Maaten2008VisualizingDU} maps in \textit{supplementary materials}.
We also demonstrate the lip synchronization improvement in Fig.~\ref{fig:abl_2d_3d_id_remove} (b) and the video on our project page. Quantitative metrics on GRID and collected dataset also validate its effectiveness as shown in Tab.~\ref{tab_abl}.

\noindent
\textbf{Evaluation of Completion-Based Generation.} We evaluate the effects of the proposed completion-based generation that benefits from jointly training on data of different people. As shown in Tab.~\ref{tab_gen_abl}, jointly training completion-based generators outperform separately training person-specific generators with much fewer network parameters when the number of speakers increases, regardless of the time length of the training data.

\begin{table}[h!]
\caption{\textbf{Training data size and training style.} PSNR/SSIM of different amount of training data and training style. The ``Joint'' means jointly training one generator for all speakers and ``Split'' means separately training multiple person-specific generators. One generation network contains 75 million parameters and $N$($N=4$ in the table) is the speaker number.}
\begin{center}
\resizebox{1.0\linewidth}{!}{
\small
\begin{tabular}{c|c|c|c|c}
\Xhline{1.2pt}
\multirow{2}{*}{Time}& 2 mins & 5 mins & 10 mins & 15 mins \\
& PSNR/SSIM & PSNR/SSIM & PSNR/SSIM & PSNR/SSIM \\ 
\Xhline{1.2pt}
Split ($N\times$ 75M) & 29.621/0.868 & 29.338/0.875 & 29.487/0.849 & 29.650/0.876 \\
\hline
\textbf{Joint (75M)} & \textbf{30.421/0.886} & \textbf{30.664/0.888} & \textbf{30.787/0.892} & \textbf{31.072/0.897} \\
\Xhline{1.2pt}
\end{tabular}
\label{tab_gen_abl}
}
\end{center}
\end{table}

\subsection{User Study}
\label{section_user_study}
To quantitatively evaluate the visual quality of generated portrait videos, following~\cite{Fried2019TextbasedEO}, we conduct a web-based user study involving 100 participants on the collected dataset. The study includes 3 generated video clips for each of the 7 cameras and for each of the 4 speakers, hence a total of 84 video clips. Similarly, we also collect $84$ ground truth video clips and mix them up with the generated video clips to perform the user study. We separately calculate the study results of the generated and ground truth video clips. 

In the user study, all the $84\times 2=168$ video clips are randomly shown to the participants and they are asked to evaluate its realism by evaluating if the clips are real on a likert scale of 1-5 (5-absolutely real, 4-real, 3-hard to judge, 2-fake, 1-absolutely fake)~\cite{Fried2019TextbasedEO}. As shown in Tab.~\ref{tab_user_study}, the generated and the ground truth video clips are rated as ``real''(score 4 and 5) in 55.0\% and 70.1\% cases, respectively. Since humans are highly tuned to the slight audio-video misalignment and generation flaws, the user study results demonstrate that our method can generate deceptive audio-video content for large poses in most cases.

\begin{table}[h]
    \centering
    \caption{\textbf{User study.} User study results on generated and ground truth video clips for videos of 7 poses.}
    \resizebox{1.0\linewidth}{!}{
    \begin{tabular}{c|cccccc|cccccc}
    \Xhline{1.2pt}
    & \multicolumn{6}{c}{Generated Videos} & \multicolumn{6}{|c}{Ground Truth Videos}\\
    \Xhline{1.2pt}
    score & 1 & 2 & 3 & 4 & 5 & ``real'' & 1 & 2 & 3 & 4 & 5 & ``real'' \\
    \Xhline{1.2pt}
    front & 5.2 & 8.5 & 20.6 & 42.6 & 23.2 & 65.8\% & 0.6 & 12.1 & 10.1 & 29.7 & 47.5 & 77.2\% \\
    up-$30^{\circ}$ & 4.6 & 25.0 & 14.2 & 36.9 & 19.3 & 56.3\% & 0.8 & 13.5 & 13.2 & 29.1 & 43.4 & 72.5\% \\
    down-$30^{\circ}$  & 4.8 & 22.0 & 15.2 & 39.7 & 18.3 & 58.0\% & 0.9 & 13.6 & 14.1 & 30.2 & 41.3 & 71.5\% \\
    right-$30^{\circ}$  & 3.9 & 22.9 & 15.8 & 42.1 & 15.3 & 57.4\% & 1.3 & 15.6 & 14.4 & 29.2 & 39.6 & 68.8\% \\
    right-$60^{\circ}$  & 7.3 & 33.8 & 11.4 & 36.9 & 10.6 & 47.5\% & 0.8 & 17.8 & 15.1 & 29.1 & 37.2 & 66.3\% \\
    left-$30^{\circ}$  & 3.2 & 20.9 & 21.9 & 40.1 & 13.9 & 54.0\% & 1.1 & 12.8 & 16.1 & 31.5 & 38.6 & 70.1\% \\
    left-$60^{\circ}$  & 7.9 & 33.7 & 12.1 & 35.1 & 11.3 & 46.3\% & 0.7 & 17.5 & 14.1 & 27.2 & 40.5 & 67.7\% \\
    \Xhline{1.2pt}
    all poses  & 5.3 & 23.8 & 15.9 & 39.0 & 16.0 & 55.0\%& 0.9 & 14.7 & 13.9 & 29.4 & 41.2 & 70.6\% \\
    \Xhline{1.2pt}
    \end{tabular}
    \label{tab_user_study}}
\end{table}

\section{Conclusion}

In this work, we present the first end-to-end learnable audio-based video editing method. At the core of our approach is the learning from audio to expression space bypassing the highly nonlinearity of directly mapping audio source to target video. Audio ID-Removing Network and Neural Video Rendering Network are introduced to enable generation of photo-realistic videos given arbitrary targets and audio sources. Extensive experiments demonstrate the robustness of our method and the effectiveness of each pivotal component. We believe our approach is a step forward towards solving the important problem of audio-based video editing and we hope it will inspire more researches in this direction.

{\small
\bibliographystyle{ieee_fullname}
\bibliography{egbib}
}

\clearpage
\section*{Appendix}
\label{appendix}
\appendix

\section{Details of Audio-to-Expression Translation Network}
\label{sec:audio_to_exp}
\begin{figure*}[t!]
    \centering
    \includegraphics[width=1\linewidth]{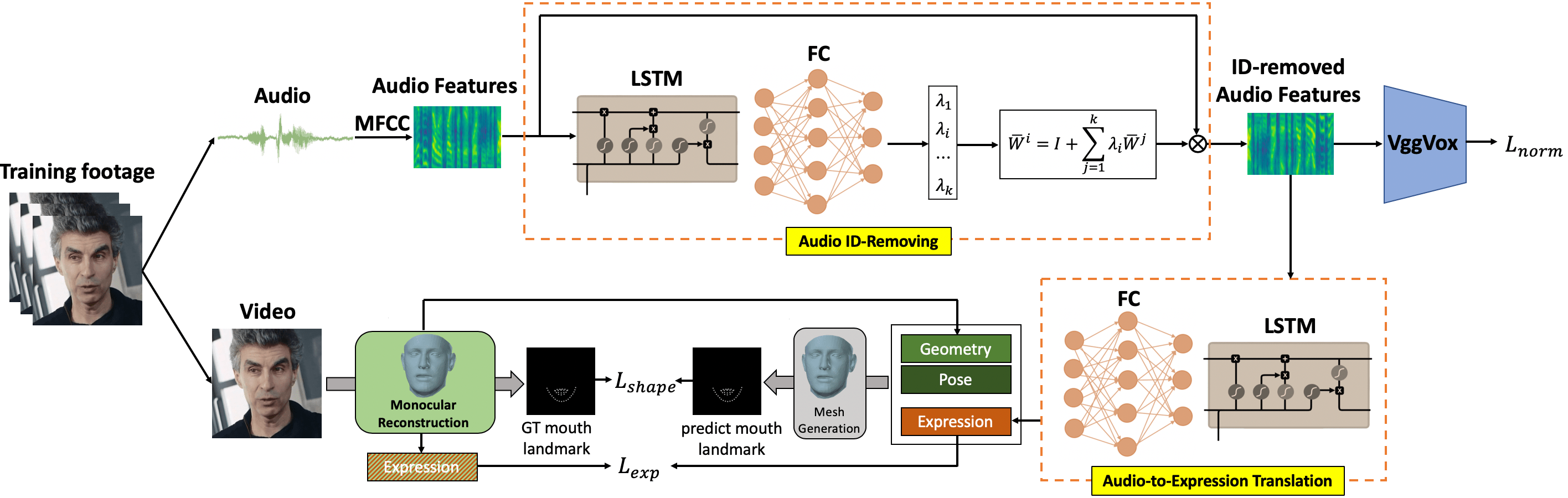}
    \caption{\textbf{Architecture of Audio-to-Expression Network.} The Audio ID-Removing Network eliminates identity information in speech audio. The Audio-to-Expression Translation Network estimate expression parameters from input audio. We constrain the predicted expression parameters and the projected 2D mouth landmark from the reconstructed facial mesh.}
    \label{fig:audio_to_exp}
\end{figure*}

The network architecture of our Audio-to-Expression Translation Network can be viewed in Figure~\ref{fig:audio_to_exp}. In the training phase, we use paired audio and video frames from the training footage as network input. Ground truth facial shape, expression and pose parameters are calculated from video frame by monocular reconstruction. From the input audio, our Audio-to-Expression Translation Network infers predicted expression parameters that are supervised by the ground truth expression parameters. The loss function is $L_{exp}=||\hat{e}-e||_2$ in Eq.~\ref{eq:l_trans}. We reconstruct facial 3D mesh by predicted expression parameters and ground truth shape parameters, then we use the ground truth pose parameters to project 2D mouth landmarks that are supervised by the ground truth 2D mouth landmarks. The loss function is $L_{shape}=||\hat{l}-l||_2$ in Eq.~\ref{eq:l_trans}. The ground truth 2D mouth landmarks are projected in a similar way where ground truth expression parameters are used. In the testing phase, the predicted expression parameters from the source audio together with ground truth shape and pose parameters from target video are used to estimate 2D mouth landmarks. The embedded mouth identity and head pose of estimated 2D mouth landmarks are the same as those of the target video while the mouth movement in accord with the source audio.

\begin{equation}
L_{trans}=L_{exp}+L_{shape} = ||\hat{e}-e||_2+||\hat{l}-l||_2
\label{eq:l_trans}
\end{equation}

\section{Temporal Flicker Removal Algorithm}
\label{sec:deflicker}
In our approach, talking face video is generated frame by frame and temporal information in video frames is only concerned in landmark estimation. During testing, we find the generated talking face videos demonstrate acceptable frame continuity even in the circumstance that the video temporal flicker removal algorithm is not applied. It is due to that audio series clips used to generate time-adjacent frames contain vast overlap in time. The remaining temporal flicker in the video can be attributed to two reasons: 1) The inferred mouth and jawline landmarks contain slight jitter. 2) Appearance flicker, especially color flicker exists in the video frames generated by the inpainting network. Based on the above analysis, our flicker removal algorithm contains two parts: mouth landmark motion smoothing and face appearance deflicker. Algorithm \ref{ldmk_smooth} demonstrates the mouth landmark motion smooth algorithm.

\vspace{-2mm}
\begin{algorithm}[H]
    \centering
    \caption{Mouth Landmark Smoothing Algorithm}
    \label{ldmk_smooth}
    \begin{algorithmic}[0]
    \Require
    $l_{t-1}$: mouth and jawline landmarks at time $t-1$;\\
    $l_{t}$: mouth and jawline landmarks at time $t$;\\
    $d_{th}$: mouth movement distance threshold, $s$: mouth movement smooth strength;
    \Ensure
    $l_{t}^{'}$: smoothed mouth and jawline landmarks at time $t$;
    \State get mouth center position $c_t$ at time $t$ from $l_t$
    \State get mouth center position $c_{t-1}$ at time $t-1$ from $l_{t-1}$
    \If {$||c_{t}-c_{t-1}||_2 > d_{th}$}
        \State $l_{t}^{'}=l_{t}$
    \Else
        \State $\alpha = \exp(-s||c_{t}-c_{t-1}||_2)$
        \State $l_{t}^{'}=\alpha l_{t-1}+(1-\alpha)l_{t}$
    \EndIf
    \State \Return $l_{t}^{'}$
\end{algorithmic}
\end{algorithm}
\vspace{-3mm}

The appearance deflicker algorithm is modified from \cite{Bonneel2015blind}. We take mouth movement into consideration. If the mouth does not move, then the color flicker is more obvious, and then we increase the deflicker strength. We denote the generated frame and processed frame at time $t$ as $P_t$ and $O_t$, respectively. The mouth center moving distance between time $t-1$ and $t$ is denoted as $d_t$. The processed frame at $t$ is written as:
\begin{equation}
\mathcal{F}(P_t) = \frac{4\pi^2f^2\mathcal{F}(P_t)+\lambda_t\mathcal{F}({\rm warp}(O_{t-1}))}{4\pi^2 f^2+\lambda_t}
\end{equation}
\noindent
where $\lambda_t=\exp(-d_t)$. Here, $\mathcal{F}$ is the Fourier transform and $f$ means frequency. Function ${\rm warp}(O_{t-1})$ uses optical flow from $P_{t-1}$ to $P_{t}$ to warp input frame $O_{t-1}$. Compared with \cite{Bonneel2015blind}, the weight of previous frame $\lambda_t$ is measured by the strength of mouth motion instead of global frame consistency.

\section{Other Experiments}
\label{sec:exp}
\subsection{Quantitative Comparison on GRID dataset}
Our method mainly focuses on talking face video editing, which is different from the recent methods that generate full face from input audio and reference still face image~\cite{Vondrick2016GeneratingV, Jamaludin2019YouST, Chen2018LipMG, Zhu2018HighResolutionTF}. Here we quantitatively compare our method with these methods~\cite{Vondrick2016GeneratingV, Jamaludin2019YouST, Chen2018LipMG, Zhu2018HighResolutionTF} on image generation metrics. For a fair comparison, in our method, we do not apply any post-process and we also modify the input of the inpainting network to generate the full face other than the mouth region. Specifically, the original network input tensor is stacked by 7 RGB frames and 7 heatmap gray images (from time $t-6$ to time $t$), we remove the RGB frame and heatmap gray image at time $t$ and require the network to generate the complete frame image at time $t$. Table \ref{tab_ssim_psnr} demonstrates that our approach outperforms these methods in PSNR and achieves comparable performance in SSIM.

\begin{table}[htbp]
    \centering
    \caption{SSIM, PSNR, IS and FID score comparison of our method and recent methods on GRID dataset. for fair comparison, we generate the full face and do not apply any post process.}
    \begin{tabular}{c|cccc}
    \toprule
    Method & PSNR & SSIM & IS & FID\\
    \midrule
    Vondrick~\etal~\cite{Vondrick2016GeneratingV} & 28.45 & 0.60 & - & - \\
    Jamaludin~\etal~\cite{Jamaludin2019YouST} & 29.36 & 0.74 & - & - \\
    Chen~\etal~\cite{Chen2018LipMG} & 29.89 & 0.73 & - & - \\
    \textbf{Ours} & \textbf{30.01} & \textbf{0.94} & \textbf{23.53} & \textbf{9.01} \\
    \bottomrule
    \end{tabular}
    \label{tab_ssim_psnr}
\end{table}

\subsection{Ablation Study on Temporal Flicker Removal Algorithm}
The temporal flicker removal algorithm tries to smooth the output landmark coordinates and eliminate the appearance flicker. The quantitative improvement is slight as shown in Table~\ref{tab_abl} but the temporal continuity improvement is obvious as shown in the video on our project page, especially when the mouth does not open. We demonstrate the quantitative results of our 3D parameter regression baseline, Audio ID-Removing Network and Temporal Flicker Removal Algorithm in Table~\ref{tab_abl}.

\begin{table}[htbp]
    \small
    \centering
    \vspace{-2mm}
    \caption{\textbf{ID-removing \& Deflicker quantitative comparison.} ``BL'' is the 3D parameter regression baseline; ``IR'' is ``id-removing''; ``DF'' is ``deflicker''. The metrics validate the effectiveness of the proposed components except for the deflicker algorithm. The deflicker algorithm mainly focus on removing temporal discontinuity that can be viewed in the video on our project page.}
    \vspace{-2mm}
    \resizebox{1.0\linewidth}{!}{
    \begin{tabular}{c|c|cc|cc}
    \Xhline{1.2pt}
    \multicolumn{2}{c|}{Method} & $E_{exp}$ & $E_{ldmk}$ & PSNR & SSIM \\
    \Xhline{1.2pt}
    \multirow{4}{*}{GRID} & BL & 0.84 & 3.09 & 30.88 & 0.94 \\
    & +IR & \textbf{0.65} & \textbf{2.24} & \textbf{32.23} & \textbf{0.97} \\
    & +DF & 0.84 & 3.07 & 27.71 & 0.92 \\
    & +IR+DF & \textbf{0.65} & \textbf{2.24} & 31.19 & 0.95 \\
    \Xhline{1.2pt}
    \multirow{4}{*}{Collected} & BL & 0.68 & 1.92 & 26.86 & 0.95 \\
    & +IR & \textbf{0.59} & 1.83 & \textbf{31.21} & \textbf{0.96} \\
    & +DF & 0.68 & 1.92 & 27.46 & 0.93 \\
    & +IR+DF & \textbf{0.59} & \textbf{1.82} & 29.16 & 0.95 \\
    \Xhline{1.2pt}
    \end{tabular}
    \label{tab_abl}}
    \vspace{-2mm}
\end{table}

\subsection{Audio ID-removing Effects in tSNE map}

\begin{figure}[h]
    \centering
    \includegraphics[width=1\linewidth]{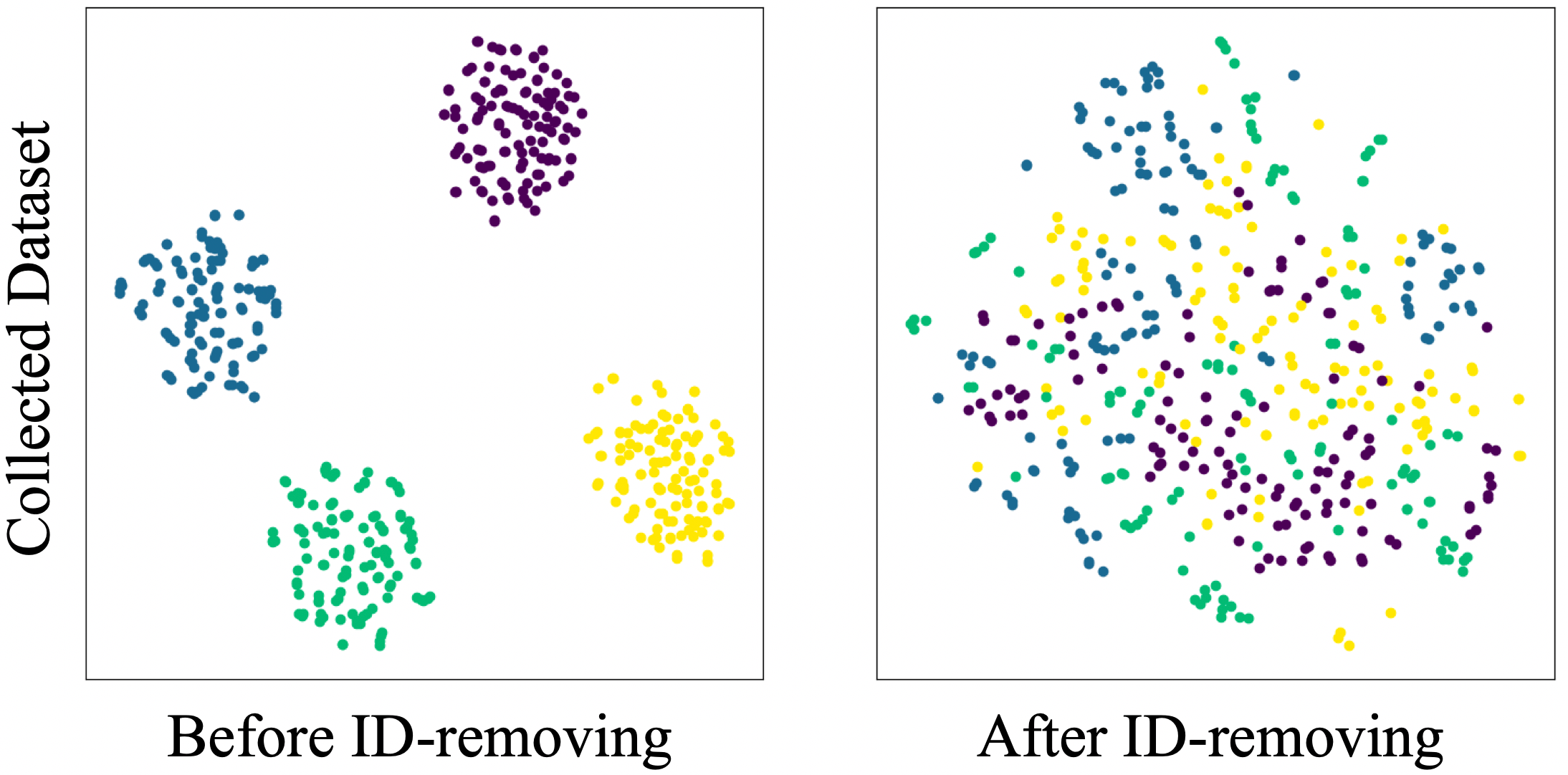}
    \caption{\textbf{tSNE before and after id-removing.} 2D visualized distributions of input MFCC and normalized MFCC. Different color represents different speaker. Our audio normalization erases the identity information embedded in the MFCC spectrum.}
    \label{fig:id_remove_tsne}
\end{figure}

The tSNE~\cite{Maaten2008VisualizingDU} maps in Figure~\ref{fig:id_remove_tsne} demonstrate the 2D visualized the distribution of the input MFCC spectrum and the identity removed MFCC spectrum produced by our Audio ID-Removing Network. We can see that the speaker identity can not be distinguished after removing identity in the MFCC spectrum. 

\section{Runtime Performance}
\label{sec:run_time}
We conduct the inference phrase on a commodity desktop computer with an NVIDIA GTX 1060 and an Intel Core i7-8700. The audio to expression network takes 17 ms per frame and the inpainting network takes 77 ms per frame. The post processes including deflicker and teeth proxy take 1.3s and 300 ms per frame respectively. The deflicker algorithm involves the calculation of optical flow that dominates the inference time. Thus, it takes about 1.7s/100ms to generate one video frame with/without the post-process on average.

\section{Limitations}
\label{sec:limit}
\textbf{Emotion}: Our method does not explicitly model facial emotion or estimate the sentiment from the input speech audio. Thus, the generated video looks unnatural if the emotion of the driving audio is different from that of the source video. This problem also appears in \cite{Suwajanakorn2017SynthesizingOL} and we leave this to future improvement.

\begin{figure}[h]
    \centering
    \includegraphics[width=1\linewidth]{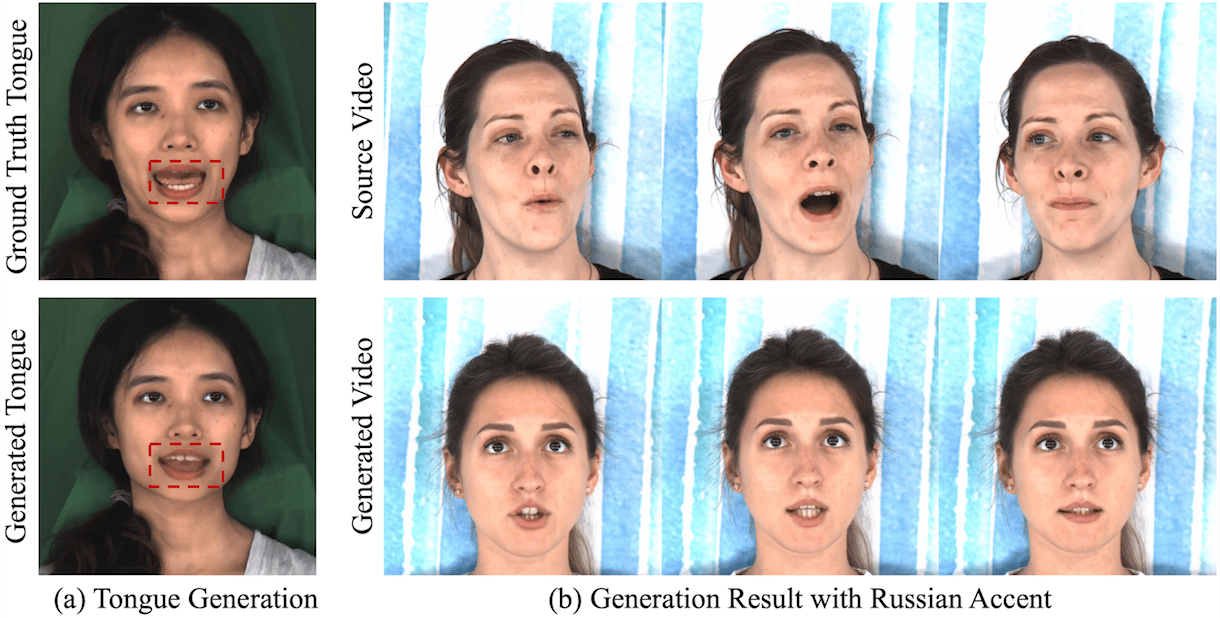}
    \caption{\textbf{Failure Cases.} (a) Poor tongue generation result on phoneme "Z" that require the use of tongue. (b) Poor lip-sync accuracy when we use normal speech audio to drive a speaker with strong Russian accent.}
    \label{fig:limitation}
\end{figure}

\textbf{Tongue}: In our method, our Neural Video Rendering Network produces lip fiducials and the teeth proxy adds the teeth high-frequency details. Our method ignores the tongue movement when some phonemes (\eg "Z" in the word "result") are pronounced. The tongue texture can not be well generated according to lip fiducials and teeth proxy as shown in Figure~\ref{fig:limitation} (a).

\textbf{Accent}: Our method performs poorly when the driving speech audio contains an accent. For example, the generated results driven by a speech with a strong Russian accent do not achieve visually satisfactory lip-sync accuracy as shown in Figure~\ref{fig:limitation} (b). We owe it to the fact that English speech with a strong accent is an outlier to our Audio-to-Expression Translation Network and we leave it to future research.

\end{document}